\documentclass[conference]{IEEEtran}
\IEEEoverridecommandlockouts
\usepackage{cite}
\usepackage{amsmath,amssymb,amsfonts}
\usepackage{algorithmic}
\usepackage{graphicx}
\usepackage{textcomp}
\usepackage{xcolor}
\usepackage{multirow}

\def\BibTeX{{\rm B\kern-.05em{\sc i\kern-.025em b}\kern-.08em
    T\kern-.1667em\lower.7ex\hbox{E}\kern-.125emX}}
    
\begin{document}

\title{FastFlow: AI for Fast Urban Wind Velocity Prediction}

\author{\IEEEauthorblockN{Shi Jer Low}
\IEEEauthorblockA{\textit{Dept of Civil \& Environ. Engineering} \\
\textit{National University of Singapore}\\
Singapore, Singapore \\
e0311219@u.nus.edu.sg}
\and
\IEEEauthorblockN{Venugopalan S. G. Raghavan}
\IEEEauthorblockA{\textit{Dept of Fluid Dynamics} \\
\textit{Institute of High Performance Computing}\\
Singapore, Singapore \\
raghavanvsg@ihpc.a-star.edu.sg}
\and
\IEEEauthorblockN{Harish Gopalan}
\IEEEauthorblockA{\textit{Dept of Fluid Dynamics} \\
\textit{Institute of High Performance Computing}\\
Singapore, Singapore \\
gopalanh@ihpc.a-star.edu.sg}
\and
\IEEEauthorblockN{Jian Cheng Wong}
\IEEEauthorblockA{\textit{Dept of Fluid Dynamics} \\
\textit{Institute of High Performance Computing}\\
Singapore, Singapore \\
wongj@ihpc.a-star.edu.sg}
\and
\IEEEauthorblockN{Justin Yeoh}
\IEEEauthorblockA{\textit{Dept of Civil and Environ. Engineering} \\
\textit{National University of Singapore}\\
Singapore, Singapore \\
Justinyeoh@nus.edu.sg}
\and
\IEEEauthorblockN{Chin Chun Ooi}
\IEEEauthorblockA{\textit{Dept of Fluid Dynamics} \\
\textit{Institute of High Performance Computing}\\
\textit{Center for Frontier AI Research}\\
Singapore, Singapore \\
ooicc@ihpc.a-star.edu.sg}
}

\maketitle

\begin{abstract}
Data-driven approaches, including deep learning, have shown great promise as surrogate models across many domains, including computer vision and natural language processing. These extend to various areas in sustainability, including  for satellite image analysis to obtain information such as land usage and extent of development. An interesting direction for which data-driven methods have not been applied much yet is in the quick quantitative evaluation of urban layouts for planning and design. In particular, urban designs typically involve complex trade-offs between multiple objectives, including limits on urban build-up and/or consideration of urban heat island effect. Hence, it can be beneficial to urban planners to have a fast surrogate model to predict urban characteristics of a hypothetical layout, e.g. pedestrian-level wind velocity, without having to run computationally expensive and time-consuming high-fidelity numerical simulations each time. This fast surrogate can then be potentially integrated into other design optimization frameworks, including generative models or other gradient-based methods. Here we present an investigation into the use of convolutional neural networks as a surrogate for urban layout characterization that is typically done via high-fidelity numerical simulation. We then further apply this model towards a first demonstration of its utility for data-driven pedestrian-level wind velocity prediction. The data set in this work comprises results from high-fidelity numerical simulations of wind velocities for a diverse set of realistic urban layouts, based on randomized samples from a real-world, highly built-up urban city. We then provide prediction results obtained from the neural network trained on this data-set, demonstrating test errors of under 0.1 m/s for previously unseen novel urban layouts. We further illustrate how this can be useful for purposes such as rapid evaluation of pedestrian wind velocity for a potential new layout. In addition, it is hoped that this data set will further inspire, facilitate and accelerate research in data-driven urban AI, even as our baseline model facilitates quantitative comparison to future, more innovative methods. 
\end{abstract}


\section{Introduction}
With the rapid increase in computing power in the past two decades, machine learning has increasingly found utility as a remarkable tool for a wide range of problems spanning image recognition and natural language processing \cite{lecun2015deep}. Excitingly, machine learning has also seen some success when applied to engineering problems such as physical simulations. This includes the prediction of flow past airfoils derived from CFD simulations \cite{thuerey2020deep,le2021surrogate}, incompressible laminar flows \cite{rao2020physics}, and other complex physics phenomena \cite{pfaff2020learning}.

Very recently, deep learning has also been applied to the field of urban characterisation, including attempts to obtain latent representations of urban layout images such as roads and building plots. These latent representations can then enable unsupervised clustering and identification of fundamental patterns in urban morphology such as building plot density and specific layout shapes \cite{moosavi2022urban,cai2021urban,chen2021classification}. This thus demonstrated the potential for deep learning to extract meaningful patterns from urban layouts, although much of the attention is naturally focused on remote sensing and sense-making from visual input such as camera and satellite images \cite{pan2020deep}. 

Simultaneously, it is often essential for urban planners to estimate how different aspects of the built environment can differ across various possible layouts. However, it is typically impossible for urban planners to obtain such data. This challenge applies both when planning for new estates and retrofitting of existing estates. Physical measurements such as solar irradiance and wind velocities are highly dependent on the surrounding developments and extremely complex and impractical to acquire. For example, it is impossible for urban planners to perform physical measurements when the planned urban layout may not exist at the time of planning. In addition, variability in weather conditions during the time of measurement can be a major confounding factor.

Thus, urban developers and planners, like the Housing Development Board (HDB) of Singapore, frequently turn to high-fidelity numerical simulations, including computational fluid dynamics (CFD) simulations, to estimate attributes such as wind profile for various possible urban layouts. These high-fidelity computer simulations, when given the correct parameters, have been shown to accurately model the physical environment \cite{blocken2015computational}. Critically, the provision of a consistent set of baseline environmental conditions allow for a fair, quantitative comparison across urban layouts. These quantitative, optimized designs are particularly important as a mitigating tool against worsening urban heat island effects under the inexorable influence of climate change \cite{luo2019challenges,manoli2019magnitude}.

Despite efforts to reduce the computation time of these simulations across decades of developments in scientific computing, these simulations remain extremely time consuming today. Due to the computational complexity of such simulations, it is extremely expensive to adopt these simulations for large scale applications. Thus, these high-fidelity computer simulations face challenges for widespread adoption in extensive and systematic optimization.

This study thus sets out to unify these different branches of development by evaluating the utility of machine learning as a computationally efficient surrogate model for important urban characteristics across diverse urban layouts. An example of these important characteristics is pedestrian-level wind velocity, which is frequently evaluated for pedestrian comfort \cite{blocken2012cfd,du2017new,du2018improving} via numerical simulations. Crucially, these measures typically involve multiple runs of time-consuming CFD simulations, comprising dozens of scenarios spanning inflow from multiple directions and/or wind velocities, further emphasizing the extremely costly nature of such evaluations.

The U-Net is a type of deep convolutional neural network architecture that has commonly been used to perform accurate image segmentation and classification. The initial success in the use of U-Net for biomedical image segmentation \cite{ronneberger2015u} has prompted further developments such as the H-DenseUNet for liver and tumour segmentation \cite{li2018h}. U-Net performs exceedingly well when the number of features within each training sample far exceeds the total number of training samples and has also been shown to perform well for multi-scale problems as is common in many physical systems such as fluid dynamics. 

As such, this study looks to adopt the U-Net, a neural network architecture that can easily be adapted towards voxelized datasets of any kind, towards the prediction of wind velocity from urban layouts. The machine learning model is trained on outdoor wind velocity fields of existing, real-world urban layouts from a heavily urbanized city, obtained through high-fidelity CFD simulations conducted in accordance with current standard guidelines for urban simulations \cite{tominaga2008aij, franke2011guide}. We then demonstrate how the use of our developed model can greatly accelerate pedestrian wind velocity evaluations for assessment of new layouts, with errors of less than 0.1 m/s. Crucially, the evaluation of new layouts can now be conducted in less than a second. This provides 3-4 orders of magnitude acceleration over the typical execution time for a comparable CFD simulation on a 24-core high-performance workstation (~3 hours).

As a proof-of-concept, the demonstration and data set here is limited to wind velocities, such as may be required for pedestrian wind comfort evaluations. Extensions to include other physics such as noise will all form important parts of a whole toolkit for the AI-assisted urban planner of the future and can be modularly added on in future work.

\section{Methods}

\subsection{Urban Layouts}\label{CutGeom}

\begin{figure}[htbp]
\begin{center}
\centerline{\includegraphics[width=0.95\linewidth]{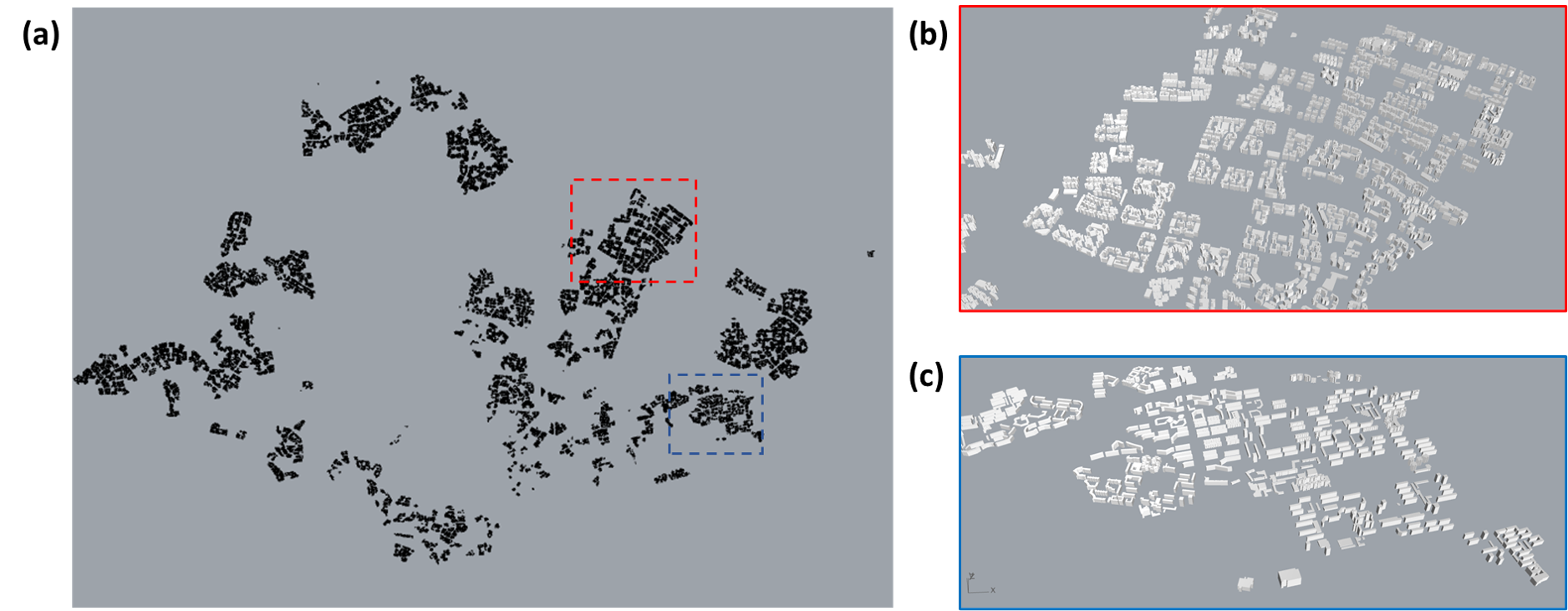}}
\caption{(a) Overall city model from which urban layouts are derived. (b)-(c) are zoomed-in views of the corresponding regions from (a) showing the variety of building geometries and heights present.}
\label{CityModel}
\end{center}
\end{figure}

The building layouts used in this project were derived from a three-dimensional city model of public housing (HDB) buildings in Singapore created in previous work by \cite{biljecki2020exploration}. The dataset comprises positions of various building geometries in the form of Level Of Detail (LOD) 1 coordinates according to the CityGML standard (Open Geospatial Consortium, 2006), as illustrated in Figure \ref{CityModel}. While the dataset also contains some attributes for each building, such as block address and year completed, the rest of this information is not actually used as inputs in the model. Nonetheless, it should be noted that the building geometries contain significant variation, having been constructed across multiple decades of development in Singapore.

\begin{figure}[htbp]
\begin{center}
\centerline{\includegraphics[width=0.85\linewidth]{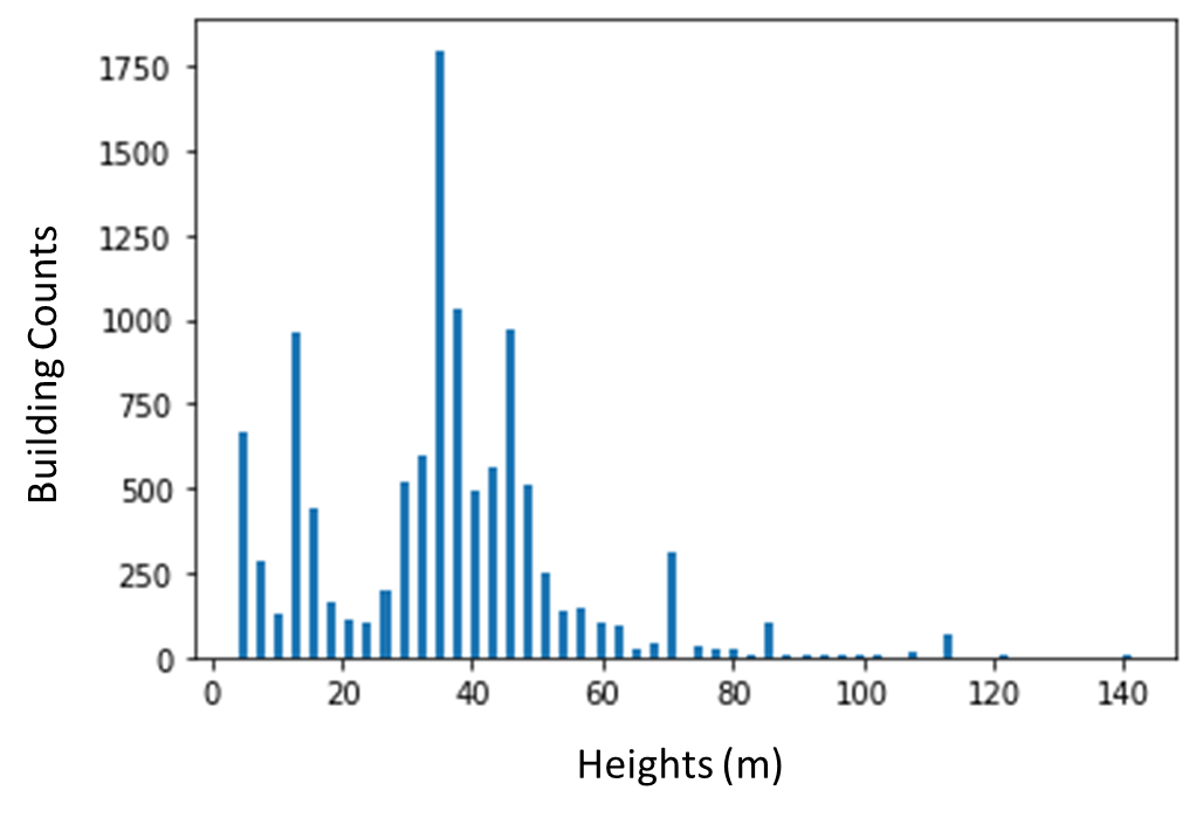}}
\caption{Histogram of building heights in the city model.}
\label{BdgHeight}
\end{center}
\end{figure}

All layouts subsequently used for the wind velocity dataset generation were randomly sampled from this starting city model. Importantly, we note that the approximately 11 thousand buildings in this city model span a spectrum of geometries and heights, as illustrated in Figure \ref{BdgHeight}.  

Briefly, buildings located wholly within a random 1 km $\times$ 1 km square area were repeatedly extracted from the entire city model to form an individual urban layout. The center of the square plot is randomly sampled from within the maximum and minimum bounds of the 3D city model while ensuring that instances where buildings intersect this square boundary are excluded to ensure complete inclusion of all building geometries. This process is repeated multiple times to form a dataset of urban layouts, which are subsequently analyzed with the aid of numerical simulations as described in Section \ref{CFD}.

\subsection{Numerical Simulations}\label{CFD}

Urban planners globally have increasingly adopted the use of high-fidelity computer simulations to simulate and understand urban characteristics such as wind ventilation. In particular, the Architecture Institute of Japan (AIJ) previously published the Guidebook for CFD Predictions of Urban Wind Environment for the simulation of wind ventilation around buildings at the pedestrian levels, which were subsequently validated with wind tunnel benchmark tests \cite{tominaga2008aij}. Other studies such as \cite{franke2011guide}, \cite{blocken2016tc} and \cite{toparlar2017tc} have focused on instituting best practices for the use of CFD for studying urban characteristics such as human thermal comfort. 

Hence, for this work, computational fluid dynamics (CFD) simulations were run in accordance with these prior guidelines for urban-scale simulations using the open source toolbox, OpenFOAM. A brief description of the modelling choices is provided in the rest of this Section.

To solve the governing Navier-Stokes equations, a modified version of the existing incompressible solver in OpenFOAM was used. While the modified solver is capable of handling temperature and buoyancy, the temperature solver was switched off for the purposes of this study. The one equation Wray-Agarwal model was used to model turbulence \cite{gopalan2018WA}. The convection term in the wind equation is discretized using a second-order upwind scheme while a bounded second-order scheme is used for the discretization of the gradient, Laplacian and face-interpolation terms. A block-coupled matrix solver with diagonal incomplete-LU (DILU) preconditioner is used for the solution of the momentum equation. All other transport equations were solved using the stabilized pre-conditioned bi-conjugate gradient method with DILU preconditioner. The pressure Poisson solver uses a preconditioned conjugate gradient method with a multi-grid preconditioner.

At the inflow, Monin-Obukhov similarity theory (MOST) is used to specify the boundary conditions. The boundary conditions for wind is calculated using the expressions given in \cite{foken2006MO}. For consistency of comparison across all cases in this work, a log-law wind profile corresponding to 2 m/s wind velocity at 15 m reference height is applied to all simulations.

To facilitate the amount of data generation desired for this study, a highly automated work-flow was developed to streamline the bulk of the simulation process. This includes meshing of each of the extracted geometries described in Section \ref{CutGeom} according to official guidelines provided by the Building and Construction Authority (BCA) of Singapore \cite{bca2016cfd}, derived based on best practices from \cite{tominaga2008aij,franke2011guide,blocken2016tc}. 

The extracted geometries are typically centered in the domain to facilitate simulation for any inlet wind direction. For example, simulation of 4 wind directions, corresponding to inflow from North, South, East and West, can be easily accomplished for each extracted geometry to enhance the dataset. The simulation of multiple wind directions for the same urban layout can be viewed as being similar to the process of data augmentation frequently done in the machine learning community.

The simulation results can then be automatically post-processed to provide values of wind speeds at discrete points at different locations and heights from the ground. The value at these points is obtained by interpolating the values at the cells, faces and vertices of the solution. 

For ease of comparison and demonstration of the model effectiveness, velocity values are extracted for 1.2m height, which is a cut-plane commonly used in urban studies, including by BCA for the evaluation of thermal comfort \cite{bca2016cfd}. This is also consistent with typical attempts to calculate pedestrian wind comfort at a single representative pedestrian height. Results are presented for this particular cut-plane in the following Sections.

\subsection{Neural Network Model}

\begin{figure}[htbp]
\begin{center}
\centerline{\includegraphics[width=0.95\linewidth]{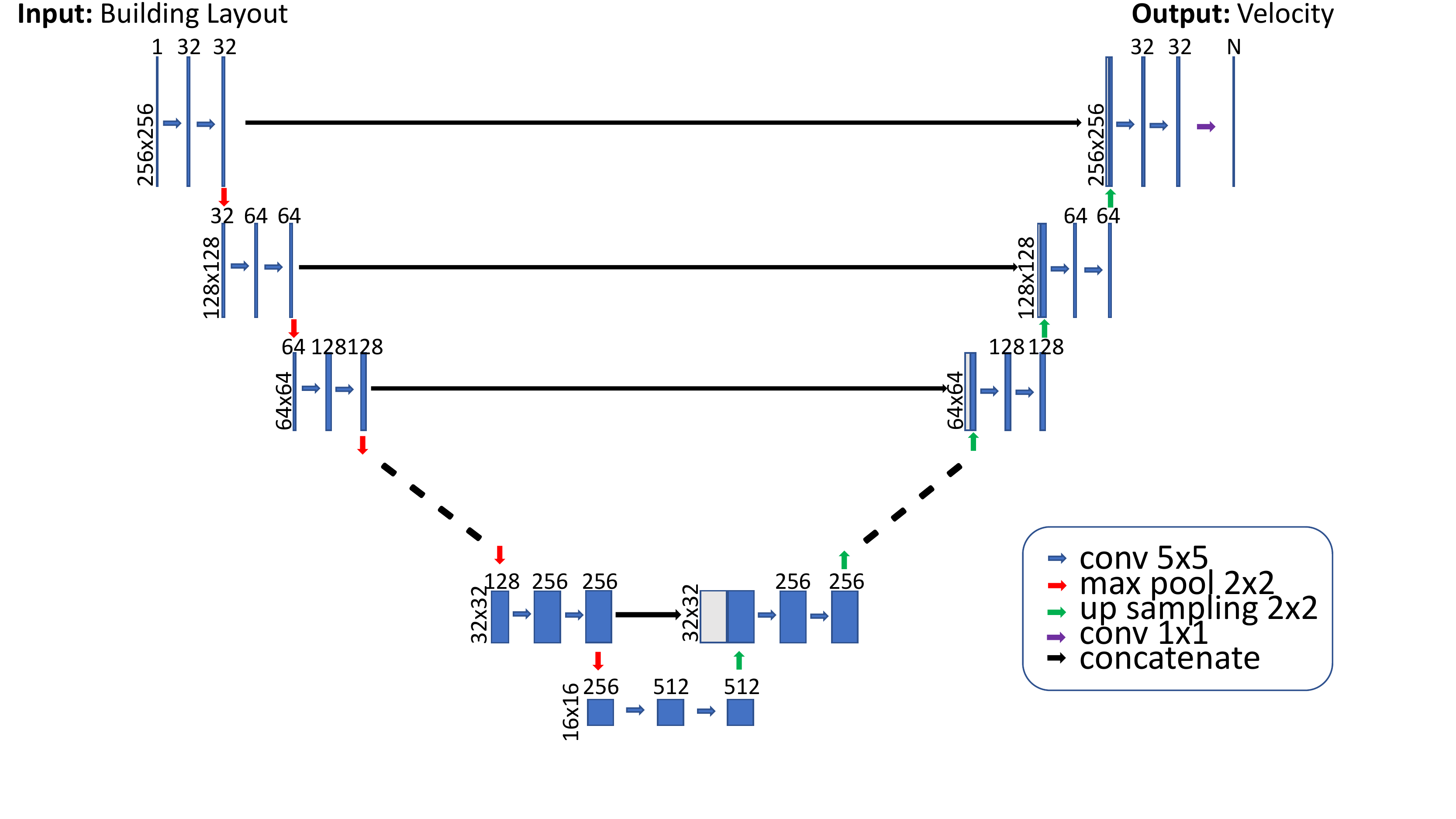}}
\caption{U-Net schematic used for this work with the accompanying layers and kernel sizes}
\label{U-Net-Sch}
\end{center}
\end{figure}

The U-Net model was created using the Keras and TensorFlow libraries. Briefly, the U-Net model contains 3 main parts. The first is a contracting phase, which extracts the latent attributes of the feature rich data samples using convolutional and max-pooling layers. Next, a bottleneck layer is used, which represents the compressed latent representation of the samples as an output of the contracting phase. Lastly, an expanding phase, where the latent representation is resized, up sampled, and concatenated with intermediate layers from the contracting phase can be used to produce predictions for the entire domain. The U-Net model architecture used in this study, with the relevant filters and convolution operators, is illustrated in Figure \ref{U-Net-Sch}.

Mean Absolute Error (MAE) was used as the loss function for training and for subsequent evaluation of the model. The loss function is defined in Eq.~\ref{data-mae}:

\begin{equation}
\label{data-mae}
    MAE = \frac{1}{N} \frac{1}{N_{X}} \sum_{i=1}^{N} \sum_{X} \Bigl | U_{i,X}^{pred} - U_{i,X}^{true} \Bigr |
\end{equation}
where $N$ is the number of layouts (data points) and $N_{X}$ is the number of pixel values per layout (corresponding to the $256 \times 256$ prediction array size). $U_{i,X}^{pred}$ and $U_{i,X}^{true}$ are the predicted and actual velocity values for each spatial point in each layout respectively. MAEs are calculated separately for each component of velocity ($u$, $v$ and $w$).

Additionally, a learning rate decay scheme was implemented to further improve optimization when training loss begins plateauing, starting with an initial learning rate of 1e-3. A batch size of 8 and kernel size of 5 was found to work better after some testing with the validation set. An L1 regularization penalty of 1e-9 was also tuned to improve model performance.

Inputs to the U-Net model are 256 $\times$ 256 arrays that encode the building height of any structure at that particular location, while the outputs are the corresponding 256 $\times$ 256 velocity arrays derived from the CFD simulations. Standard pre-processing of the input and output arrays are done to ensure that the scales of the input and output vectors are maintained at less than 1.

The entire simulation dataset comprises of 1088 different urban layouts and 4352 simulations (4 wind directions per layout) and this is split into three groups for model training (888 layouts), validation (25 layouts), and testing (25 layouts). The validation and test sets comprise simulation results for 4 wind directions for each layout, hence, the total number of cases in the validation and test set is 100 each. The code for the U-Net model is available online at Github (URL to be updated after review). However, the data-set is more than 6GB, even for a single velocity cut-plane, and is currently not uploaded online, although the authors will be willing to share upon request.

\section{Results}

\subsection{Ground Truth Wind Speeds}\label{AA}

\begin{figure}[htbp]
\begin{center}
\centerline{\includegraphics[width=0.95\linewidth]{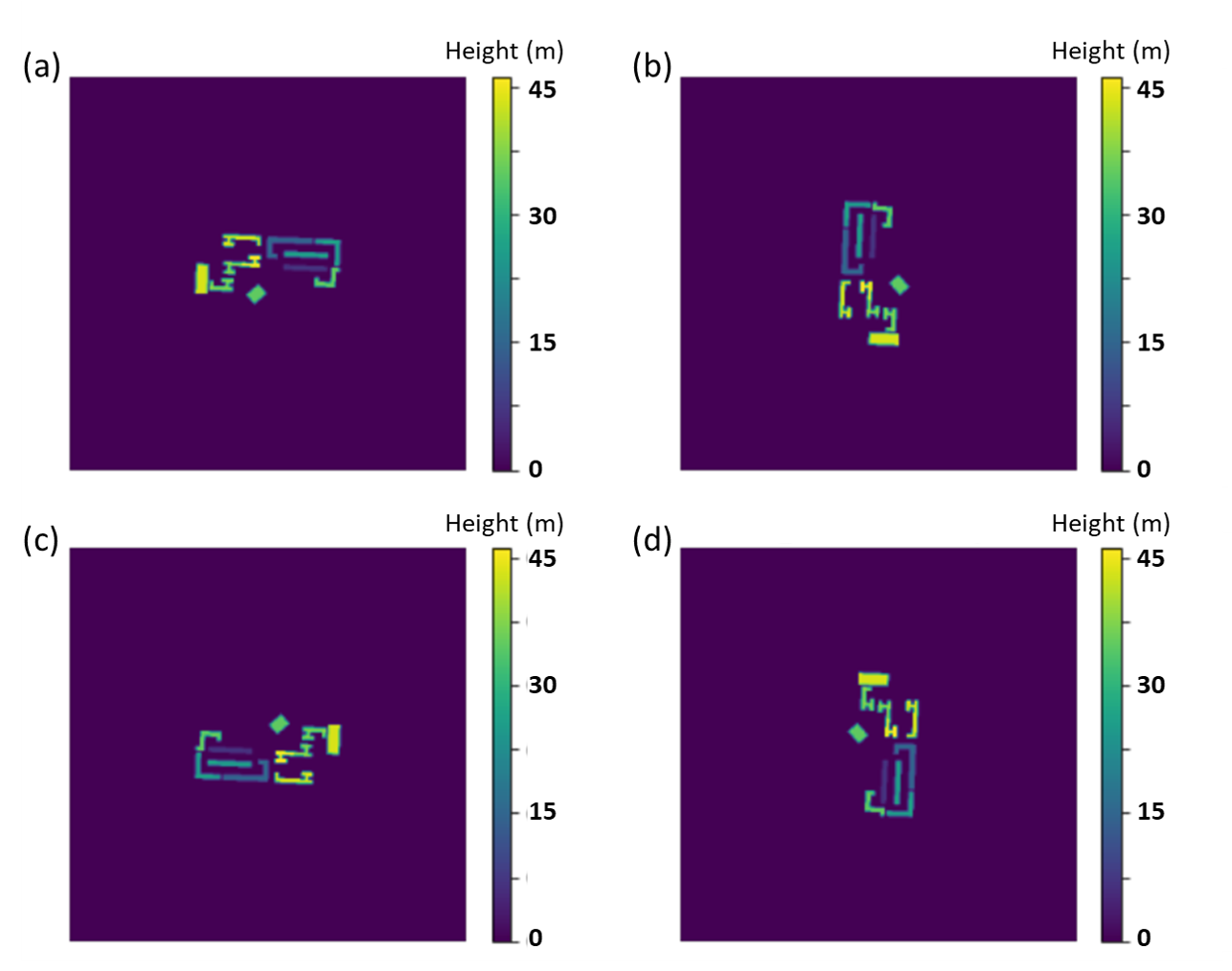}}
\caption{4 rotations of the same building layout used for CFD}
\label{Rotate4}
\end{center}
\end{figure}

Figure \ref{Rotate4} demonstrates how 4 rotations corresponding to physical inflows from different directions for the same layout are used in the creation of this data set. Hence, the data set comprises 1088 layouts with 4 simulation directions each.

\begin{figure}[htbp]
\begin{center}
\centerline{\includegraphics[width=0.95\linewidth]{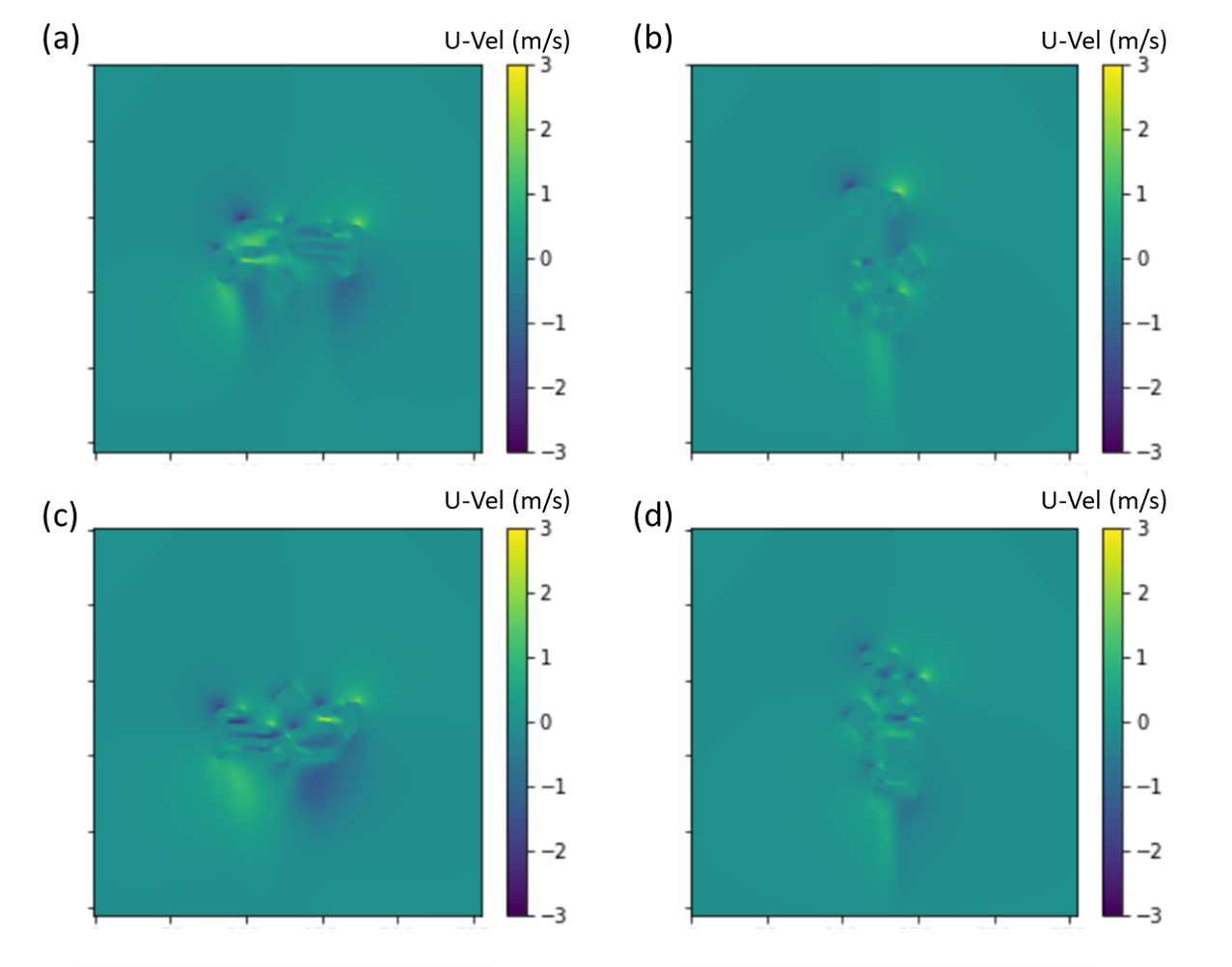}}
\caption{U-velocity contours for the 4 CFD simulations conducted for the building layout in Figure \ref{Rotate4}}
\label{4-U}
\end{center}
\end{figure}

\begin{figure}[htbp]
\begin{center}
\centerline{\includegraphics[width=0.95\linewidth]{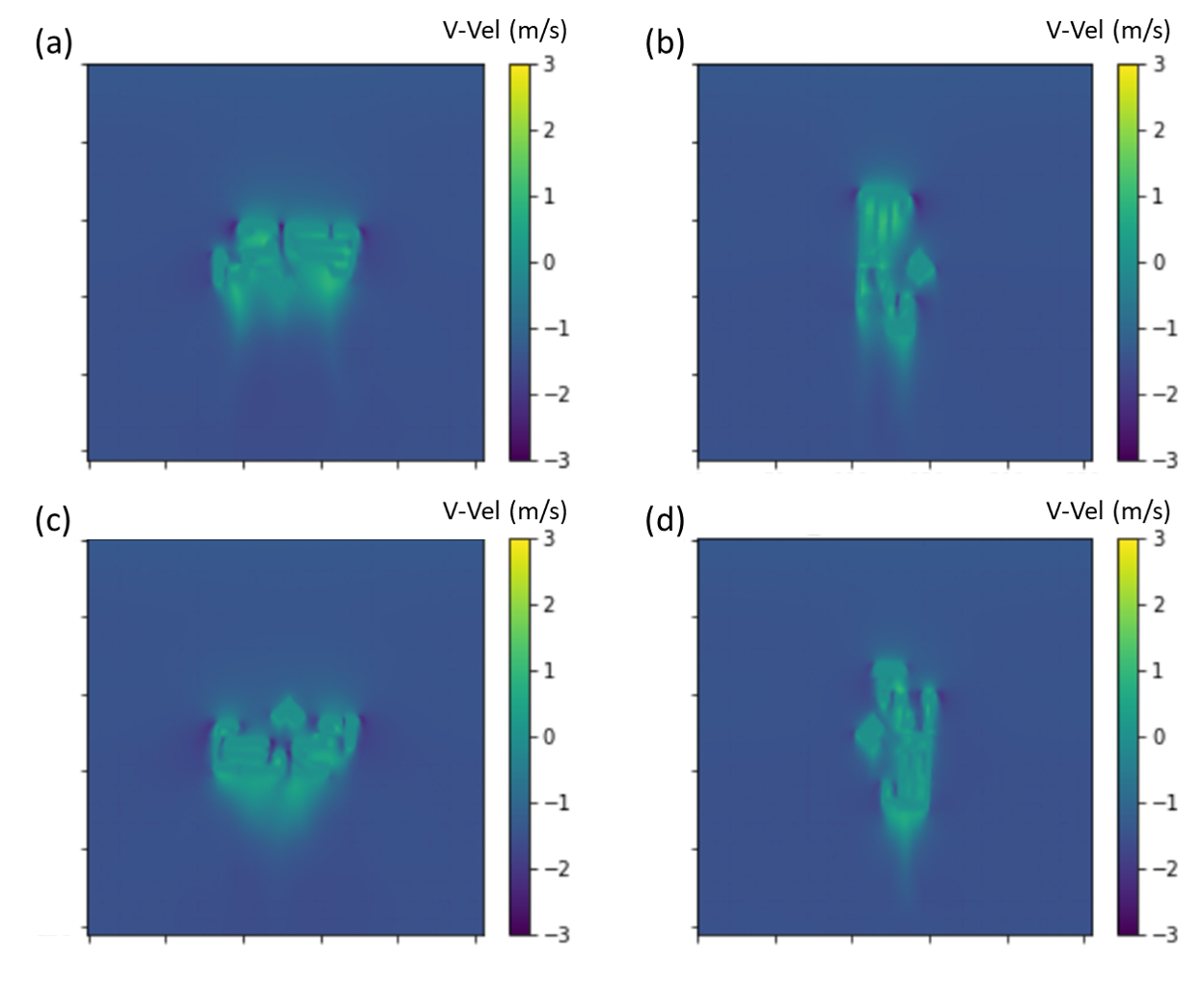}}
\caption{V-velocity contours for the 4 CFD simulations conducted for the building layout in Figure \ref{Rotate4}}
\label{4-V}
\end{center}
\end{figure}

Figures \ref{4-U} and \ref{4-V} are representative examples of the types of U-velocity and V-velocity contours that result from the high-fidelity CFD simulations conducted.

Statistics to describe the distribution of velocities from the variety of urban layouts are listed in Table \ref{tab: vel-stats}. The distributions show how U and W are centered around zero while the V-component has a bias in the downwards direction due to the provided inflow boundary condition. In addition, the W-component (vertical velocity) is substantially weaker than the horizontal components of the velocity.

\begin{table}[htbp]
\caption{Statistical descriptors of simulation velocities}
\begin{center}
\begin{tabular}{|c|c|c|c|}
\hline
\multirow{2}{*}{\textbf{Measure}} & \multicolumn{3}{|c|}{\textbf{Velocities (m/s)}} \\
\cline{2-4}
{} & {$U$} & {$V$} & {$W$} \\
\hline
{Mean} & {$-5.3 \times 10^{-4}$} & {$-1.168$} & {$-1.0 \times 10^{-4}$} \\
\hline
{Min} & {$-5.947$} & {$-7.303$} & {$-1.108$} \\
\hline
{Max} & {$5.825$} & {$2.794$} & {$0.610$} \\
\hline
{Std Dev} & {$0.299$} & {$0.602$} & {$0.015$} \\
\hline
\end{tabular}
\label{tab: vel-stats}
\end{center}
\end{table}

For consistency in model training, the simulation-derived velocity results for the different wind inflow directions are rotated to ensure that all inputs to the U-Net maintain the same wind inflow direction (in the negative V direction). This does not affect generalizability of the model as is demonstrated in Section \ref{PedWind}. A wind velocity prediction for any wind direction of interest can be conveniently obtained for any urban layout via an equivalent rotation of the urban layout. This facilitates adoption for evaluations of urban layouts across multiple wind directions, including in pedestrian wind velocity assessment.

\subsection{Neural Network Model}

U-Net models were then trained to predict the simulated U, V and W components of the velocity fields with results averaged across 3 replicates presented in Table \ref{tab: uvw-result}.

\begin{table}[htbp]
\caption{Model Test Error}
\begin{center}
\begin{tabular}{|c|c|c|c|}
\hline
\multirow{2}{*}{\textbf{Error Statistic}} & \multicolumn{3}{|c|}{\textbf{MAE (m/s)}} \\
\cline{2-4}
{} & {$U$} & {$V$} & {$W$}\\
\hline
{Mean} & {$0.0568$} & {$0.0815$} & {$0.00342$}\\
\hline
{Std} & {$0.0003$} & {$0.0002$} & {$0.00003$}\\
\hline
\end{tabular}
\label{tab: uvw-result}
\end{center}
\end{table}

\begin{figure}[htbp]
\begin{center}
\centerline{\includegraphics[width=0.9\linewidth]{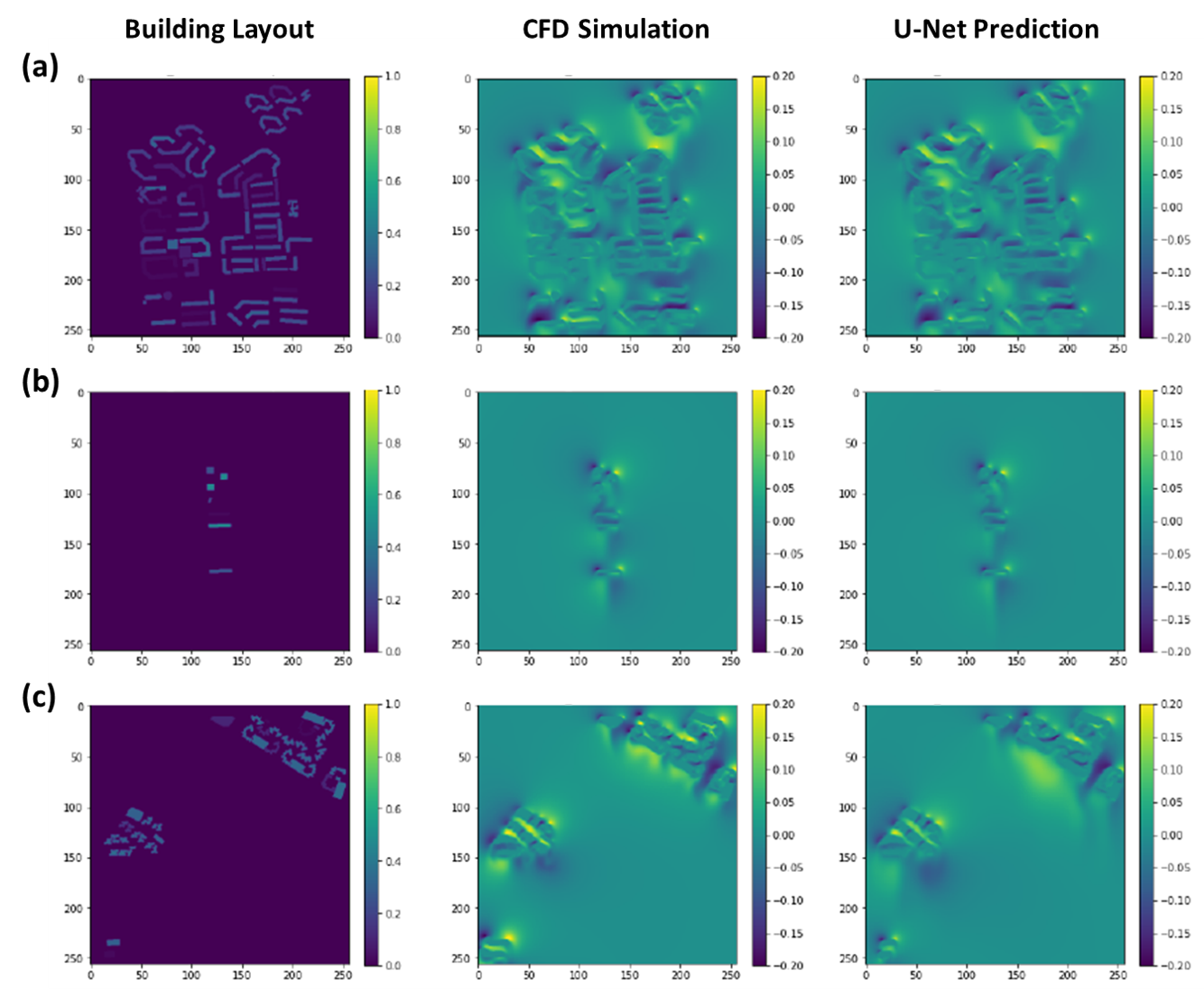}}
\caption{(a)-(c) are examples comparing the (Right) U-Net predictions for U-velocity to the (Middle) CFD ground truth for the (Left) urban layouts depicted.}
\label{U-Vel-Example}
\end{center}
\end{figure}

\begin{figure}[htbp]
\begin{center}
\centerline{\includegraphics[width=0.9\linewidth]{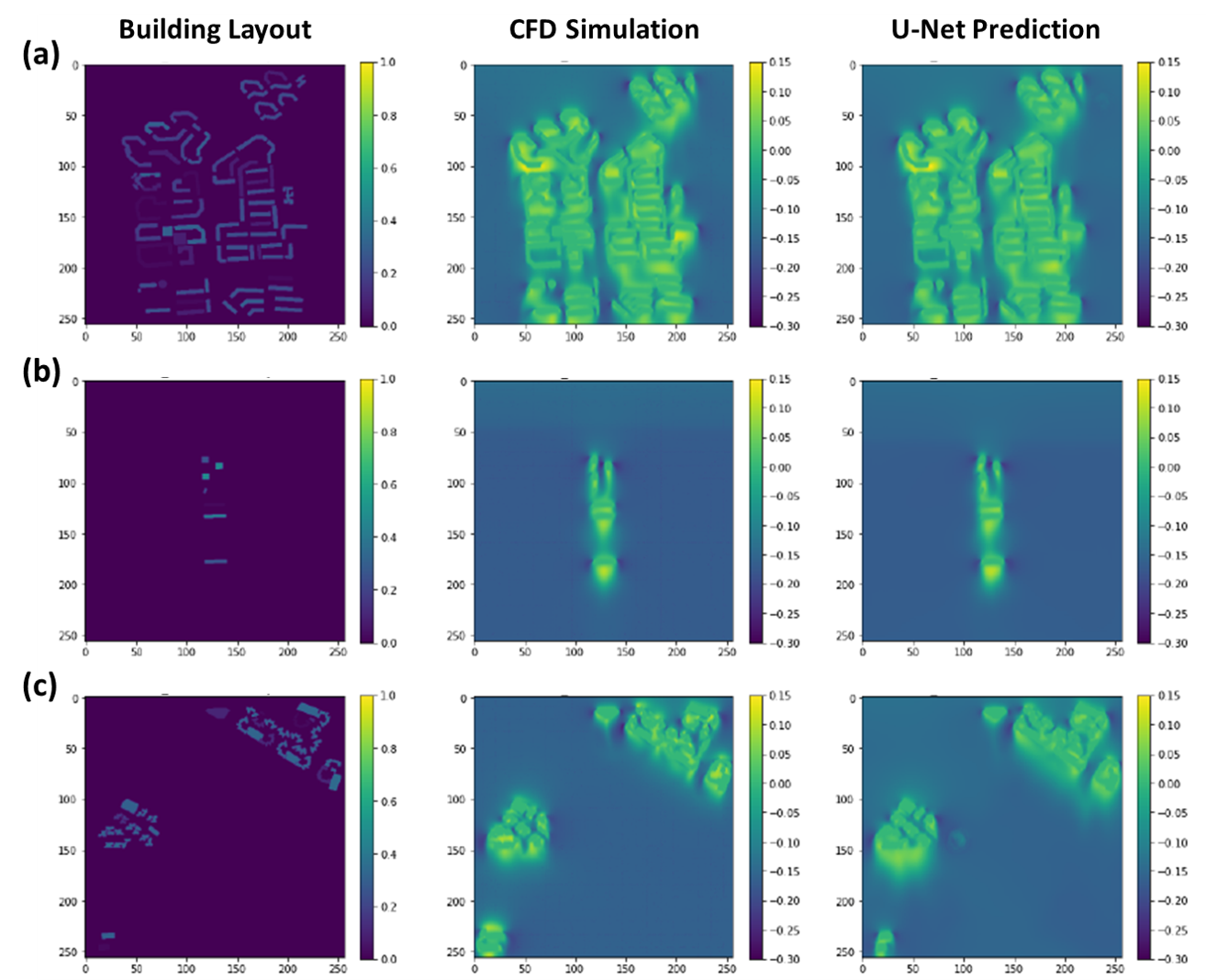}}
\caption{(a)-(c) are examples comparing the (Right) U-Net predictions for V-velocity to the (Middle) CFD ground truth for the (Left) urban layouts depicted.}
\label{V-Vel-Example}
\end{center}
\end{figure}

Representative U and V velocity predictions of 3 samples from the test dataset are provided in Figure \ref{U-Vel-Example} and Figure \ref{V-Vel-Example} respectively.

For both velocity components, the wind flow patterns present in the wakes are a lot more diffused than the ground truth values. The U-Net models also perform well in replicating the complex wind flow characteristics present between buildings of a high-dense area, although there are still minor differences in certain more complex regions.


Prediction results on the test set show that the U-Net model can predict U-component velocities with a mean absolute error of 0.0568 m/s, V-component velocities with a mean absolute error of 0.0815 m/s, and W-component velocities with a mean absolute error of 0.0034 m/s. The W-component velocity is much smaller in general, and the errors are also an order of magnitude lower, suggesting that U/V velocity components are dominant. In addition, model performance is generally quite consistent, with little variation across replicate runs.

Overall, the U-Net produces errors of under 0.1 m/s in magnitude, which is extremely useful, especially as errors in field measurements themselves inherently can have errors on the order of or greater than 0.05 - 0.1 m/s. Taken relative to the individual dataset velocity standard deviations, we note that these are at least 5-10x smaller, suggesting that the U-Net model has indeed successfully learnt the wind flow patterns present across various urban layouts. 

In addition, it has been shown across many other domains such as computer vision that the model performance depends critically on the quantity of training data. Hence, it is anticipated that an increased amount of training data will directly lead to improvements in performance, especially as the U-Net is trained with more examples. Indeed, the results in Table \ref{tab: data-dep} show that a doubling of the amount of provided data lead to an approximate decrease in MAE of about 0.01 m/s. The results also clearly show that the current dataset is still not comprehensive enough for an asymptotic trend to have developed. Interestingly, we note that there appears to be a log-linear trend in the relationship between improvements in model performance and the amount of data provided for this particular range of dataset size. This trend was previously reported in work by \cite{thuerey2020deep}, using a similar U-Net type architecture, suggesting that certain similar dataset and model characteristics could be at play.

\begin{table}[htbp]
\caption{Model Dependence on Dataset Size}
\begin{center}
\begin{tabular}{|c|c|c|c|}
\hline
\multirow{2}{*}{\textbf{Dataset Size}} & \multicolumn{2}{|c|}{\textbf{MAE (m/s)}} \\
\cline{2-3}
{} & {$U$} & {$V$} \\
\hline
{100} & {$0.0865 \pm 0.0002$} & {$0.1183 \pm 0.0001$} \\
\hline
{200} & {$0.0783 \pm 0.0002$} & {$0.1041 \pm 0.0002$} \\
\hline
{400} & {$0.0699 \pm 0.0001$} & {$0.0985 \pm 0.0002$} \\
\hline
{888} & {$0.0568 \pm 0.0003$} & {$0.0815 \pm 0.0002$} \\
\hline
\end{tabular}
\label{tab: data-dep}
\end{center}
\end{table}

\begin{figure}[htbp]
\begin{center}
\centerline{\includegraphics[width=0.9\linewidth]{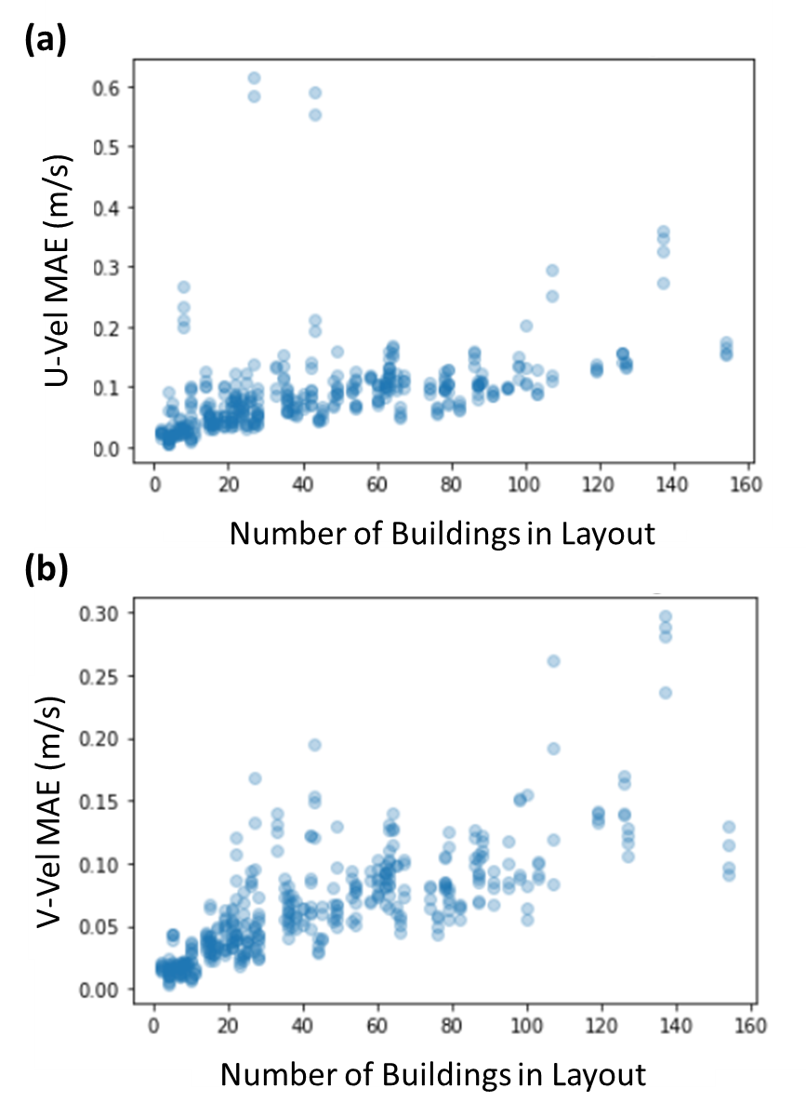}}
\caption{Scatterplots of test MAE against number of buildings in the layout for both (a) U-velocity and (b) V-velocity}
\label{BdgDensityErr}
\end{center}
\end{figure}

Upon further analysis, it was observed that the mean absolute error for each layout in the test set is correlated to the number of buildings in each layout, as per Figure \ref{BdgDensityErr}. This observation is physically consistent as well, as increased number of buildings can greatly increase the complexity of urban layouts, making features such as wakes harder to predict accurately. 

Given that the current dataset is not comprehensive, it is hypothesized that the active selection of more dense layouts will naturally be more informative, and lead to better model performance under conditions of data sparsity. Hence, we further select the top 100 and 200 densest layouts for inclusion in a reduced training dataset, and evaluate the model's performance relative to the randomly sampled dataset in Table \ref{tab: dense-sel}. This is anticipated to provide insights into how to better select training samples for data generation a priori. 

\begin{table}[htbp]
\caption{Active Sampling by Building Density}
\begin{center}
\begin{tabular}{|c|c|c|c|}
\hline
\multirow{2}{*}{\textbf{Dataset}} & \multicolumn{2}{|c|}{\textbf{MAE (m/s)}} \\
\cline{2-3}
{} & {$U$} & {$V$} \\
\hline
{100-Random} & {$0.0865 \pm 0.0002$} & {$0.1183 \pm 0.0001$} \\
\hline
{100-Dense} & {$0.0898 \pm 0.0001$} & {$0.1398 \pm 0.0002$} \\
\hline
{200-Random} & {$0.0783 \pm 0.0002$} & {$0.1041 \pm 0.0002$} \\
\hline
{200-Dense} & {$0.0750 \pm 0.0003$} & {$0.1135 \pm 0.0002$} \\
\hline
\end{tabular}
\label{tab: dense-sel}
\end{center}
\end{table}

Interestingly, the active selection of denser, seemingly more informative urban layouts for training did not appear to benefit the model performance. It is anticipated that other more efficient data selection techniques to enable such active sampling could be essential here.

\subsection{Pedestrian-Level Wind Comfort}\label{PedWind}

There have been numerous prior work in literature describing the use of high-fidelity CFD for the evaluation and improvement of pedestrian-level wind comfort. In particular, we note that CFD is typically used to evaluate the pedestrian level wind velocity for the location of interest multiple times, in an attempt to properly characterize the location across the range of wind velocities and directions it may encounter across the entire year. 

\begin{figure}[htbp]
\begin{center}
\centerline{\includegraphics[width=0.9\linewidth]{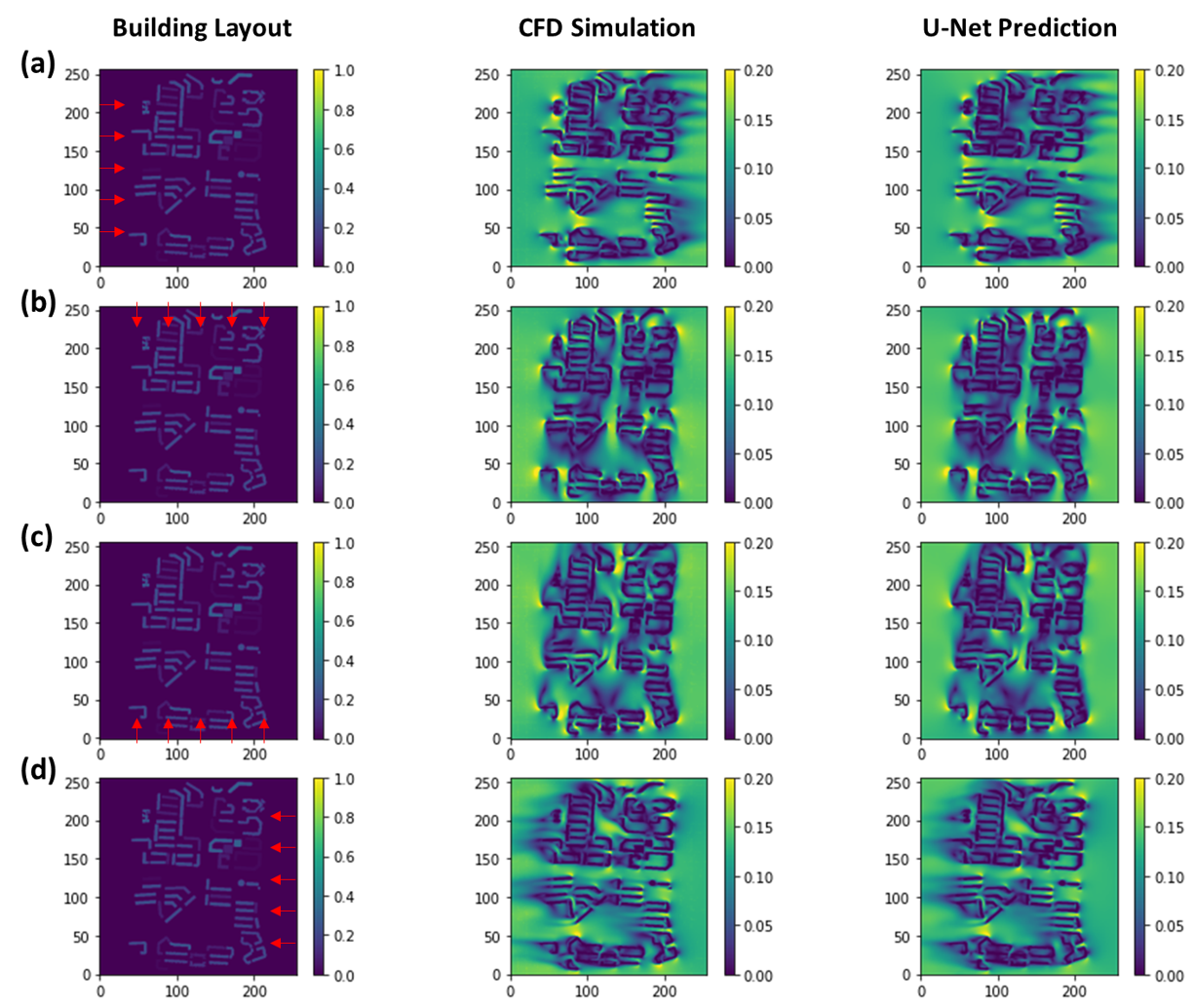}}
\caption{(a)-(d) are examples showing the (Right) U-Net predictions for velocity magnitude relative to (Middle) CFD ground truth for the (Left) wind directions depicted with the red arrows. This is for a single complex urban layout as depicted.}
\label{VelMag4Dir}
\end{center}
\end{figure}

Hence, we further demonstrate in this work how the U-Net model can be used to predict the wind velocity magnitude for a pedestrian-level cut-plane across 4 wind directions. Contours of the predicted velocities are provided in Figure \ref{VelMag4Dir} and demonstrate good correspondence between CFD ground truth and U-Net model prediction.

\begin{figure}[htbp]
\begin{center}
\centerline{\includegraphics[width=0.9\linewidth]{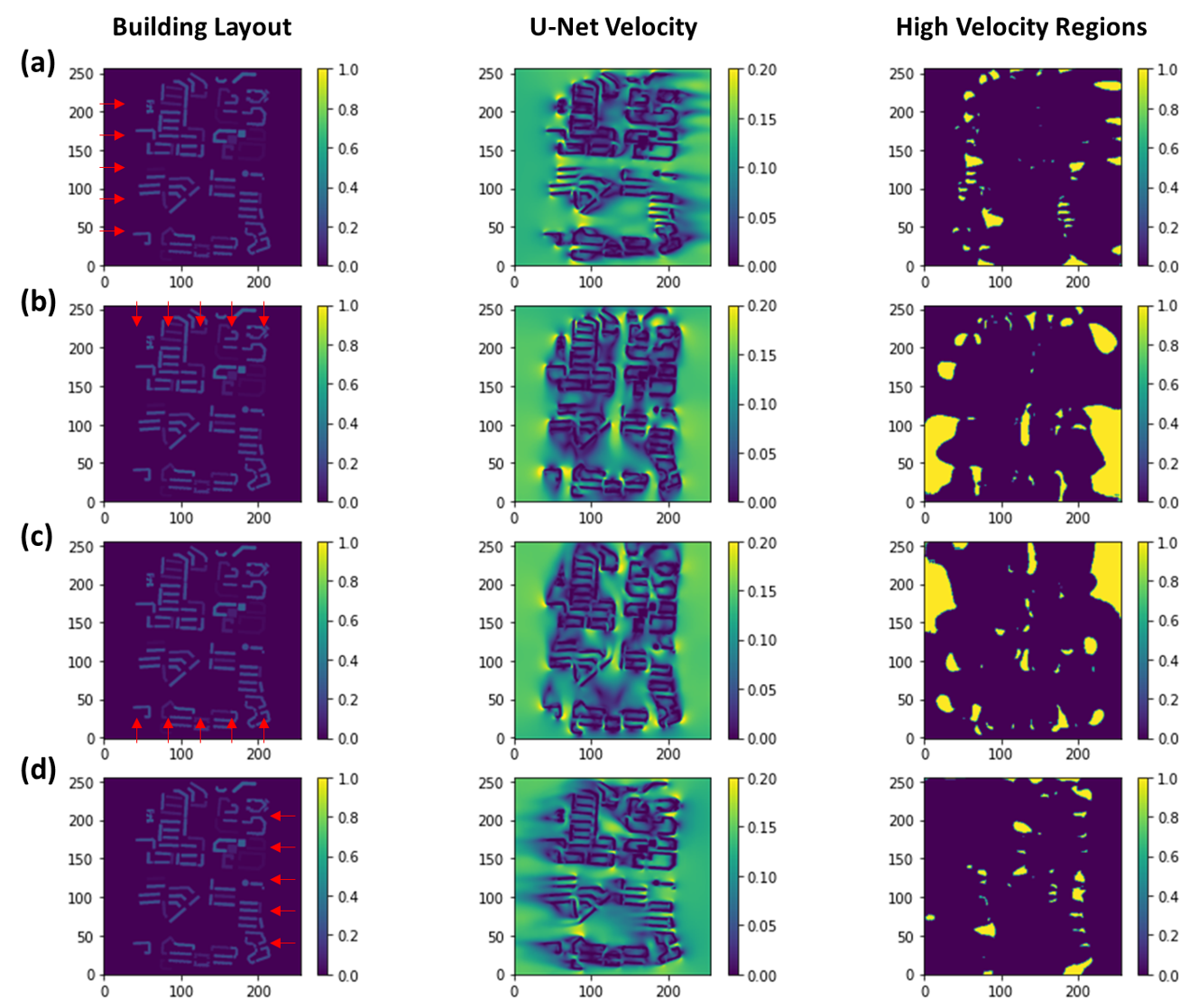}}
\caption{(a)-(d) are examples showing the (Middle) U-Net predictions for velocity magnitude and the (Right) corresponding high velocity (> 1.5 m/s [Yellow]) regions within the domain for the (Left) wind directions depicted with the red arrows. This is for a single complex urban layout as depicted.}
\label{VelMag4Criteria}
\end{center}
\end{figure}

While several standards exist, as a proof-of-concept demonstration, we use a guideline proposed in earlier work that suggests minimum velocities of 1.5 m/s is a desirable threshold for low wind conditions in high-density urban areas \cite{ng2009policies}. Hence, we further convert the layout into a mask showing the regions within this layout that can successfully reach 1.5 m/s in Figure \ref{VelMag4Criteria}. 

In this particular instance, we note that the layout is particularly dense, and much of the internal domain is unable to reach 1.5 m/s. In addition, for brevity, we only show how this can be done for 4 wind directions, although additional simple rotations of the building geometries will suffice to create new inputs to the U-Net model for other wind directions as required. 

More importantly, this proof-of-concept demonstration is a clear example of how the U-Net can be used to greatly facilitate analysis and evaluation of urban layouts in a timely and informative manner.

\section{Discussion}

The U-Net models achieved good accuracy as approximation models, in terms of their mean absolute test errors. The models are able to predict the shape and directions of wind flow around buildings on new layouts. The ability of the model to make predictions for new layouts in the test set points to its generalizability. However, in areas with high building density, distinct wind flow patterns becomes diffused. Buildings at the edge of the layouts may also cause higher errors in some instances.

More importantly, the U-Net models produced in this study are computationally inexpensive and produce results in less than a second, whereas a typical CFD simulation can take more than 3-4 hrs on a 24-core processor, which is a difference in time taken of more than 1000x. This advantage is compounded when the typical need for evaluation across multiple wind directions is taken into account, such as is demonstrated in the case of pedestrian wind velocity assessment. Hence, the speed and accuracy of the U-Net models make them suitable for use as a preliminary analysis tool. 

In a study where training data is sparse because they are expensive to generate, samples should be chosen carefully to justify the computational cost of each sample. Since the urban layouts used as training samples in this study were picked at random, there may be a lot of redundancies within the training data. Also, with a relatively small dataset, the samples chosen may also not be representative of edge cases. Thus, in the future, active sampling methods, such as via unsupervised clustering of the urban layout images, can be utilized to generate more informative data points. Interestingly, active selection of samples by the most intuitive characteristic, building density, did not appear to result in substantially better model performance. This further highlights the potential need for auto-encoders and unsupervised learning to extract other latent attributes to improve the sampling under a restricted computational budget.

In addition, we note that the use of a fairly standard U-Net model was presented in this work to demonstrate the effectiveness of a simple, baseline model for the fast, data-driven prediction of wind flows across various urban layouts. We further note that the use of additional, innovative model architectures, including physics-informed models, can potentially lead to even better performances.

It is also worth noting that we believe this to be among the first demonstrations of the use of AI as a rapid surrogate model in urban layout evaluation. The promising results here, in particular, suggest that further extension towards more complex physics, including temperature effects and buoyancy, may also be possible, leading to the creation of a quick and easy-to-use AI-enabled urban planning toolkit. Naturally, some of the considerations grappled with by urban planners may also need to be addressed in the context of AI methods. For example, urban codes typically explicitly model buildings for an approximate $1 \times 1 km^{2}$ area. However, the choice of appropriate input layer size for an AI model might be different and can be investigated in future work.

\section{Conclusion}

In this work, we provide proof-of-concept results that a simple U-Net neural network can be trained on a relatively small data-set to provide fast, fairly accurate predictions of wind velocities for new, previously unseen urban layouts with errors of under 0.1 m/s. This demonstrates a simple baseline which will facilitate direct comparison to future, more innovative methods. In particular, it is hoped that this dataset will further inspire, facilitate and accelerate research in data-driven urban AI.



\bibliography{mybibfile}

\end{document}